# Analysis of Blood Report Images Using General-Purpose Vision-Language Models

Nadia Bakhsheshi
*dept. Electrical Engineering*
Sharif University of Technology
Tehran, Iran
nadia.bakhsheshi1380@gamil.com

Hamid Beigy
*dept. Computer Engineering*
Sharif University of Technology
Tehran, Iran
beigy@sharif.edu

*Abstract*—The reliable analysis of blood reports is important for health knowledge, but individuals often struggle with interpretation, leading to anxiety and overlooked issues. We look for the potential of general-purpose Vision-Language Models (VLMs) to address this challenge by automatically analyzing blood report images. We conduct a comparative evaluation of three VLMs—Qwen-VL-Max, Gemini 2.5 Pro, and Llama 4 Maverick—determining their performance on a dataset of 100 diverse blood report images. Each model was prompted with clinically relevant questions adapted to each blood report. The answers were then processed using Sentence-BERT to compare and evaluate how closely the models responded.

The findings suggest that general-purpose VLMs are a practical and promising technology for developing patient-facing tools for preliminary blood report analysis. Their ability to provide clear interpretations directly from images can improve health literacy and reduce the limitations to understanding complex medical information. This work establishes a foundation for the future development of reliable and accessible AI-assisted healthcare applications. While results are encouraging, they should be interpreted cautiously given the limited dataset size.

*Keywords—Vision-Language Models, VLM, Health Care AI, Blood Report Analysis, Multimodal AI*

## I. Introduction

We observed that many patients brought their lab reports without understanding basic markers such as hemoglobin or cholesterol. Also, in many countries, people do not have immediate access to doctors (getting an appointment is very difficult), so they are genuinely concerned about their health issues.

On the other hand, since technology has advanced rapidly and many APIs are now freely accessible, we want to explore whether these APIs can be used, how accurate their responses are, and whether they share any similarities in their functionality.
This motivated us to test whether general-purpose VLMs could help to have a coherent interpretation.

Recent advancements in artificial intelligence have given rise to Vision-Language Models (VLMs), which enhance visual understanding and natural language processing (NLP). By integrating a visual encoder with a large language model (LLM), VLMs can interpret multimodal inputs (images and text) and generate textual responses [1], [2]. They have shown strong performance across diverse domains, including visual question answering and document understanding. However, the performance of general-purpose VLMs on specific tasks such as interpreting images of blood test reports remains largely unexplored.

We aim to bridge this gap by conducting a comparative analysis of three general-purpose VLMs—Qwen-VL-Max by Alibaba Cloud, Llama 4 maverick by Meta, and Gemini 2.5 Pro by Google—in interpreting patient blood report images. Our methodology involves prompting each model with a specific question, relevant prompts (e.g., for blood report number 1, take question 1which is what the patient asked), then getting the model's answer, and using Sentence-BERT [3] to generate embeddings for all model outputs and calculating their cosine similarity. Our findings offer valuable insights into the comparative performance of VLMs as accessible tools for preliminary blood report analysis.

The remainder of the paper is organized as follows. In the related work section, we highlight specific points from previous papers that inspired our work. The next section, background and preliminaries, provides a brief explanation of the basic concepts to help readers better understand the paper. Then, in the methodology section, we present and explain our proposed method in detail. In the results section, we show the outcomes of our experiments. The discussion section explores potential directions for future research inspired by our findings. Finally, the conclusion summarizes the paper and offers closing remarks.

## II. Related Work

Vision-Language Models (VLMs) have proven to be a powerful tool in the medical field, particularly for tasks that require the joint understanding of visual and textual data. The diagnostic process for complex diseases like cancer often relies on synthesizing information from multiple sources, including radiology images, clinical notes, and laboratory reports, such as blood reports. VLMs are uniquely suited for this integration, as they can process both imaging and text inputs to generate comprehensive analyses [2].

In the medical domain, VLMs have shown promising results in diagnostic support and report generation. A remarkable example is MammoVLM [4], a generative model specifically designed for mammography. It integrates a visual encoder trained on mammographic images with a language model to generate detailed diagnostic reports. Fine-tuned on large-scale, domain-specific datasets, MammoVLM shows strong performance in lesion detection, classification, and report generation. This paper asked 48 patients about the questions that came to their minds—questions related to mammograms. Our work, which focuses on the kinds of questions people might ask, was inspired by theirs. Their dataset is also very comprehensive, consisting of more than 30,000 mammography reports. Their model performs very well on unseen data.

Similarly, XrayGPT[5] is designed for the interactive analysis of chest radiographs. It combines a frozen medical visual encoder (MedCLIP) with a fine-tuned large language model (Vicuna) through a linear projector. Trained on over 217,000 curated radiology summaries, it excels at generating concise, clinically relevant responses to open-ended questions about X-rays. This two-stage training process—first on broad image-text pairs and then on radiology-specific data—ensures both robustness and domain alignment.

Another valuable example is models like RadVLM[6], which adapts general-purpose VLM architectures for radiology. Trained on paired image-report data from corpora like MIMIC-CXR. RadVLM uses incompatible learning to align image features with textual descriptions, enhancing its performance in report generation, abnormality detection, and clinical Q&A. This work is fine-tuned on rare categories, which enhances its performance. They also use real patient questions, making their work more user-friendly.

There is also "Generative AI in Academic Writing: A Comparison of DeepSeek, Qwen, ChatGPT, Gemini, Llama, Mistral, and Gemma"[7] investigation into how this general VLM compares the academic writing capabilities of several large language models (LLMs), and how it generates and paraphrases texts using these models based on academic papers related to Digital Twin and Healthcare. In this work, they check whether these models produce academically acceptable, original content. Also, other LLMs are used to check semantic similarities.

Despite these advancements, few studies have explored the performance of general-purpose VLMs on structured clinical documents such as blood reports. Blood reports are typically considered as images of structured documents containing tables and quantitative data, often without direct interpretive text. This gap motivates our investigation into how general-purpose VLMs—originally designed for broad multimodal tasks—perform in interpreting blood report images and whether they can be adapted for reliable clinical support.

III. BACKGROUND AND PRELIMINARIES

Vision-Language Models (VLMs) are multimodal AI systems that combine a vision encoder with a large language model (LLM), enabling the LLM to process and interpret visual inputs such as images alongside textual prompts. This architecture allows VLMs to perform tasks that require both visual recognition and linguistic reasoning, including image captioning, visual question answering, and document understanding [1], [2].

Most VLMs follow a three-component architecture:

1. Vision Encoder: Typically based on transformer models (e.g., CLIP), trained on large-scale image-text datasets to associate visual features with semantic meaning [1].
2. Projector: A set of layers that transforms the output of the vision encoder into a format compatible with the LLM, often represented as image tokens [1].
3. Large Language Model (LLM): Processes the combined input of text and image tokens to generate context-aware textual responses [1].

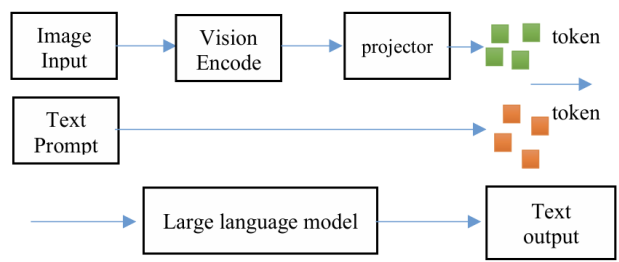

**Fig. 1.** Standard architecture of a Vision-Language Model (VLM). The Vision Encoder processes the input image into embeddings, which are then projected into the language model's space by the Projector module. The Large Language Model (LLM) merges these visual tokens with text tokens from the text prompt to generate a textual response.

IV. METHODOLOGY

This section outlines the methodology used to evaluate the performance of vision-language models on blood report images. We begin by describing the dataset and preprocessing steps, followed by an overview of the selected models and API configuration. Next, we detail the prompting strategy used to simulate realistic patient queries. Finally, we explain the evaluation procedure, including the use of Sentence-BERT and cosine similarity to compare model outputs.

*4.1 Dataset*

A dataset of 426 images of blood reports was sourced from the publicly available "Medical Lab Report Dataset" on Kaggle [8]. Although the Kaggle dataset provided 426 images, some were low quality or duplicated. After manual cleaning, we selected 100. The reports originated from various institutions, ensuring a diversity of formats and layouts. All images were pre-processed, and any personally identifiable information (PII) was already removed by the data source provider.

*4.2 Models and API*

Three VLMs were selected for this study:
- **Qwen-VL-Max** (by Alibaba Cloud)
- **Gemini 2.5 Pro** (by Google)
- **LLaMA 4 Maverick** (by Meta)

The models were executed using Python in a Google Colab environment. The source code for data processing, API calls, and analysis is available in our public GitHub repository.

API calls were configured with consistent parameters across all models to ensure deterministic and comparable outputs with a maximum token set of 1024. Each blood report image was encoded into a base64 string, which is a standard format for transmitting binary data (like images) over web-based APIs, and included in the multimodal prompt request as specified by each provider's documentation.

*4.3 Prompting Strategy*

To simulate a realistic patient interaction, we employed a custom prompt designed for each image. For each of the 100 blood report images, a unique prompt was generated by a human reviewer who asked the question a patient would likely ask after examining their specific report (e.g., "Is my A/G ratio in normal range?"). This approach ensured that each model was queried with a highly relevant and realistic question for each individual report. Crucially, for a given report image, all three models received the exact same prompt, ensuring a fair comparison.

*4.4 Evaluation*

To provide a quantitative measure of agreement between the model outputs, we calculate semantic similarity scores. The textual responses from each model were converted into high-dimensional vector embeddings using the Sentence-BERT library, specifically the 'all-MiniLM-L6-v2' model [3].

Subsequently, cosine similarity was calculated between the embedding vectors of each pair of models (Qwen-Gemini, Qwen-LLaMA, Gemini-LLaMA) for every report. Cosine similarity measures the cosine of the angle between two vectors, providing a score between -1 and 1, where 1 indicates identical semantic meaning. The formula is:

$$similarity = cos(\theta) = \frac{A \cdot B}{\|A\|\|B\|}$$

Where A and B are the embedding vectors of two model outputs.

## V. RESULTS

This section presents the findings from the semantic similarity analysis, which measured the consensus between the model outputs.

Figure 2 presents sample responses from the three VLMs to the question "Is my A/G ratio in the normal range?"

The pairwise cosine similarity was calculated for each of the 100 blood report images. The average semantic similarity scores between each model pair are reported in Table 2 and visualized in Fig. 3.

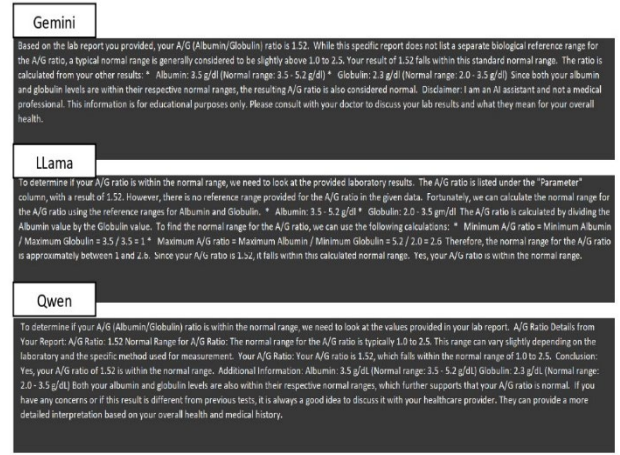

Fig. 2. In this figure, the image shows the answer to the question "Is my A/G ratio in the normal range?" The first thing is that the answers are very similar to each other and show a lot of similarity, which is also confirmed by the cosine similarity reported in Table I.

TABLE I

| Qwen-Gemini | Qwen-LLaMA | Gemini-LLaMA | Average Similarity |
|---|---|---|---|
| 0.929695189 | 0.93600297 | 0.879383028 | 0.915027082 |

TABLE II

| Models pair | Average similarity |
|---|---|
| Qwen vs. Gemini | 0.797 |
| Qwen vs. Llama | 0.816 |
| Gemini vs. Llama | 0.794 |
| Overall average | 0.803 |

The results indicate a high degree of semantic alignment between the outputs of all three models, with an overall average similarity score of 0.803. The highest agreement was observed between the outputs of Qwen and Llama.

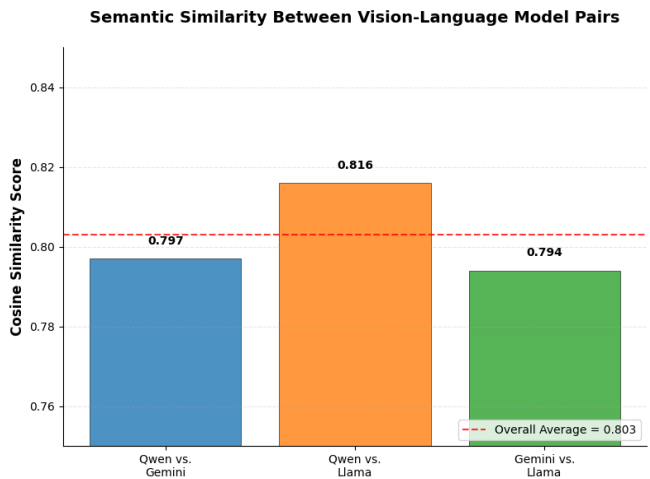

**Fig. 3.** Cosine similarity scores. The similarity between Qwen and Gemini (mean score: 0.803) suggests that both models produced consistent and informative responses.

## VI. DISCUSSIONS

*6.1 Result Interpretation*

We aimed to evaluate the capability of general-purpose Vision-Language Models (VLMs) to analyze blood report images. The high overall semantic similarity score (0.803) indicates a strong agreement in the clinical interpretations generated by Qwen, Gemini, and Llama. This suggests that, despite their different architectures and training data, these models have learned a shared understanding of how to process and reason about structured medical data presented in images. The slightly higher agreement between Qwen and Llama (0.816) warrants further investigation.

*6.2 Future Work*

- Expanding the evaluation to a wider variety of medical documents (e.g., radiology reports, mammography reports)
- Incorporate more specialists to define a metric for the rate of clinically reliable and valuable answers, and to have more reliable answers.

## VII. CONCLUSION

Our results demonstrated a high degree of semantic alignment in the models' outputs, with an overall average cosine similarity score of 0.803. This strong agreement indicates that these VLMs have developed a shared understanding of how to process and reason about structured clinical data presented in images, specifically for blood report images.

The primary meaning of this finding is that general-purpose VLMs can serve as a possible foundation for preliminary patient-facing healthcare applications, such as automated report explanation tools. Their ability to generate coherent analyses directly from image inputs reduces the need for complex, specialized pipelines involving separate OCR and NLP models. This could significantly lower the technological limitations to creating accessible tools for improving public health literacy.